\pdfoutput=1

\documentclass[11pt]{article}

\usepackage[preprint]{acl}

\usepackage{times}
\usepackage{latexsym}

\usepackage[T1]{fontenc}

\usepackage[utf8]{inputenc}

\usepackage{microtype}

\usepackage{inconsolata}

\usepackage{graphicx}

\usepackage{pifont} 
\usepackage{soul} 

\definecolor{lightblue}{rgb}{0.8,0.85,1}
\definecolor{lightgreen}{rgb}{0.85,1,0.85}
\definecolor{lightred}{rgb}{1,0.85,0.85}
\definecolor{lightpurple}{rgb}{0.9,0.85,1}
\definecolor{lightgrey}{rgb}{0.9,0.9,0.9}

\newcommand{\hlblue}[1]{\sethlcolor{lightblue}\hl{#1}}
\newcommand{\hlgreen}[1]{\sethlcolor{lightgreen}\hl{#1}}
\newcommand{\hlred}[1]{\sethlcolor{lightred}\hl{#1}}
\newcommand{\hlpurple}[1]{\sethlcolor{lightpurple}\hl{#1}}
\newcommand{\hlgrey}[1]{\sethlcolor{lightgrey}\hl{#1}}

\title{Can Language Models Rival Mathematics Students? Evaluating Mathematical Reasoning through Textual Manipulation and Human Experiments}

\author{
  \textbf{Andrii Nikolaiev\textsuperscript{1}},
  \textbf{Yiannos Stathopoulos\textsuperscript{2}},
  \textbf{Simone Teufel\textsuperscript{2}} \\
  \textsuperscript{1}Taras Shevchenko National University of Kyiv \\
  \textsuperscript{2}University of Cambridge \\
  {
    \texttt{nikolaev@knu.ua},
    \texttt{\{{yiannos.stathopoulos, simone.teufel}\}@cl.cam.ac.uk}
  }
}

\begin{document}
\maketitle
\begin{abstract}
In this paper we look at the ability of recent large language models (LLMs) at solving mathematical problems in combinatorics. We compare models LLaMA-2, LLaMA-3.1, GPT-4, and Mixtral against each other and against human pupils and undergraduates with prior experience in mathematical olympiads. To facilitate these comparisons we introduce the \emph{Combi-Puzzles} dataset, which contains 125 problem variants based on 25 combinatorial reasoning problems. Each problem is presented in one of five distinct forms, created by systematically manipulating the problem statements through adversarial additions, numeric parameter changes, and linguistic obfuscation. Our variations preserve the mathematical core and are designed to measure the generalisability of LLM problem-solving abilities, while also increasing confidence that problems are submitted to LLMs in forms that have not been seen as training instances. We found that a model based on GPT-4 outperformed all other models in producing correct responses, and performed significantly better in the mathematical variation of the problems than humans. We also found that modifications to problem statements significantly impact the LLM's performance, while human performance remains unaffected.
\end{abstract}

\section{Introduction}
Mathematical problem-solving tasks are often presented as textual statements, requiring solutions that are logical, coherent, and culminate in a correct final answer. The task of solving mathematical word problems (MWP) has been a topic of interest in the fields of computational linguistics and artificial intelligence (AI) since the 1960s.

Solving MWP as an area of research has seen a resurgence due partly to research questions afforded by new AI models, assessing the reasoning abilities of large language models (LLMs). Similar to human thinking patterns, efficient application of models usually requires accurately parsing text to identify relevant information, discard irrelevant details, and understand relationships between entities. Additionally, it requires creative reasoning to identify relevant arguments and steps, technical proficiency for performing routine calculations, and critical reasoning to examine and adjust the steps of an argument as mentioned by \citet{bubeck2023sparksartificialgeneralintelligence}. Applying AI to solving advanced mathematical problems is now an open-challenge task for researchers, with challenges like AIMO Prize\footnote{\url{https://aimoprize.com/}.} being a notable example.

Recent studies demonstrated that LLMs like GPT-4 exhibit human-like reasoning \cite{wei2022emergent, openai2024gpt4technicalreport} and can tackle various natural language reasoning tasks. \citet{kojima2023largelanguagemodelszeroshot} found that GPT-4 performs well at reasoning tasks without further fine-tuning, but only experimented with problems requiring few reasoning steps to solve. Later, \citet{collins2023evaluatinglanguagemodelsmathematics} created a set of novel problems, presented them to GPT-4 and observed that the model has a tendency to over-rely on plausibly memorised examples or patterns. Therefore, whether the reasoning demonstrated by LLMs generalises is an open question that requires further investigation.

This study examines the current abilities of established LLMs in mathematical reasoning using combinatorial problems with comparisons to the abilities of human participants. We present a fresh dataset and specialised methodology aimed at evaluating the generalisation of reasoning abilities in LLMs and investigate three questions:
\begin{enumerate}
    \item Which LLM is the best at answering combinatorial reasoning problems?
    \item Are humans better at reasoning on these problems than the best LLM?
    \item Which problem variations affect LLMs and humans the most?
\end{enumerate}

\section{Background and Related work}
Most studies on language model evaluation in mathematical reasoning typically focus on mathematical word problems (MWPs). Datasets for this task are usually structured as mathematical statements and queries of mathematics-related concepts. Among the most widely utilised datasets are GSM8K \cite{cobbe2021trainingverifierssolvemath}, SVAMP \cite{patel2021nlpmodelsreallyable} with a focus on arithmetic reasoning, MMLU \cite{hendrycks2021measuringmassivemultitasklanguage} -- multiple choice question answering, and MATH \cite{hendrycks2021measuringmathematicalproblemsolving} -- word problem-solving. Some instances of combinatorial problems are included in the listed datasets. However, the emergence of LLMs has highlighted several issues:
\begin{itemize}
    \item Common datasets lack balanced difficulty levels, featuring either simple arithmetic or challenging olympiad-style problems.
    \item Some datasets include test data from downstream tasks in the training data, causing pattern memorisation instead.
    \item Few studies provide human evaluation of datasets, crucial for validating problem-solving abilities.
\end{itemize}

To address these issues, several new datasets were introduced in recent years.

\paragraph{Problems complexity.} To enhance the difficulty level, the aforementioned datasets have been extended, and new datasets including challenging problems were introduced.

\citet{zheng2022minif2fcrosssystembenchmarkformal} presented miniF2F, a dataset of formal Olympiad-level mathematics problem statements intended to provide a unified cross-system benchmark for neural theorem proving. \citet{frieder2023mathematicalcapabilitieschatgpt} opted to create the GHOSTS dataset of graduate-level mathematics. \citet{wang2024mmluprorobustchallengingmultitask} introduced the MMLU-Pro dataset, including data from TheoremQA, featuring high-quality, human-annotated questions that necessitate the application of theorems for resolution; and SciBench, which includes advanced science questions derived from college exams, ensuring the inclusion of curriculum-aligned questions. In the PuzzleBench dataset, \citet{mittal2024puzzlebenchllmssolvechallenging} shortlisted computationally challenging problems by manually scanning Wikipedia for diverse puzzles and NP-hard algorithmic problems. Additionally, the authors experimented with altering the configurations of certain problems to curate the complexity level of the tasks.

\paragraph{Data contamination.} Authors utilised GSM8k to replicate the questions in the dataset while withholding their corresponding parts and successfully identified contamination of the dataset within GPT-4 when guided instruction was used \cite{golchin2024timetravelllmstracing}.

\citet{zhang2024carefulexaminationlargelanguage} highlighted contamination issues with existing datasets and mirrored the style and complexity of the GSM8k data to create their own dataset GSK1k using fresh problems contributed by human annotators. The models tested showed significantly worse results on the alternative problem set, which effectively highlights the issue of LLM's memorisation of data. In the most recent research done \citet{mirzadeh2024gsmsymbolicunderstandinglimitationsmathematical}, authors presented an updated GSM-symbolic dataset, an improved benchmark created from symbolic templates that allow for the generation of a diverse set of questions, including changing names, numbers or both, injection of irrelevant numerical information into the problem statements. The findings reveal that LLMs exhibit noticeable variance when responding to different instantiations of the same problem question.

\paragraph{Human evaluation.} \citet{collins2023evaluatinglanguagemodelsmathematics} conducted the first experiment with humans and LLMs in the domain of mathematics. The authors assessed the helpfulness of LLMs as mathematical assistants through direct interactions with undergraduate students and professors, and found notable instances of divergence between correctness and perceived helpfulness in LLM generations. In the work by \citet{zheng2023judgingllmasajudgemtbenchchatbot}, authors found that LLMs show limitations in grading basic math problems which it is capable of solving. The test model, GPT-4, was able to solve the problem when asked separately, but it was misled by the provided answers, ultimately resulting in incorrect judgment. In the work by \citet{bubeck2023sparksartificialgeneralintelligence}, the authors prompted the GPT-4 model with mathematical statements previously not seen online. From the experiments, the model demonstrates a high level of ability in choosing the right argument or path towards the solution but frequently fails to conduct correct reasoning due to arithmetic mistakes.

Our goal is to evaluate the reasoning ability of established LLMs and draw comparisons with human subjects. We opted to create our dataset -- the \emph{Combi-Puzzles} dataset of combinatorial problems -- with fresh problems addressing all issues above. First, we included dataset problems of various difficulty levels to challenge all parties participating in the study. Second, fresh problem statements are needed in order for our evaluation to be unbiased -- it gives us more confidence that good LLM performance on our dataset is not due to dataset memorisation. Third, by creating our own dataset we introduce controlled alterations to problem statements designed to measure the ability of LLMs and participants to identify and reason about the mathematical problem underlying each textual statement or core.

\begin{table}
\centering
\small
\begin{tabular}{|p{0.38\linewidth}|p{0.31\linewidth}|p{0.13\linewidth}|}
\hline
\textbf{Problem statement} & \textbf{Combinatorial answer} & \textbf{Num. answer} \\
\hline
Lia has 2 apples, 3 bananas and 2 oranges. For the upcoming week, she wants to eat one fruit every day. How many ways are there to do it? & \[ P(2,3,2) = \frac{7!}{2!3!2!} \] & \[ 210 \] \\
\hline
\end{tabular}
\caption{A combinatorial problem example provided with an answer.}
\label{tab:puzzle_example}
\end{table}


\begin{table*}
\centering \small
\begin{tabular}{|p{2.5cm}|p{12cm}|}
\hline
\textbf{Problem form} & \textbf{Problem example} \\ \hline
Common & 3 girls found 9 white pearls. How many distinct ways are there to divide all pearls between girls? It is not necessary that all girls get pearls. \\ \hline
Mathematical & In an urn there are 3 balls numbered 1 to 3. You draw 9 times with replacement. How many distinct sets of balls are there? \\ \hline
Adversarial & 3 girls found 9 white pearls. \hlred{All girls are professional free divers and can hold their breath from 8 to 10 minutes.} How many distinct ways are there to divide all pearls between girls? It is not necessary that all girls get pearls. \\ \hline
Parameterisation & \hlgreen{13 girls} found \hlgreen{54 white pearls}. How many distinct ways are there to divide all pearls between girls? It is not necessary that all girls get pearls. \\ \hline
Linguistic obfuscation & 3 pirates enter a frigate that has just surrendered to them. They know that there are nine bars of solid gold on board. According to pirate law, any pirate who finds some loot on board a commandeered ship can keep it. They swarm out into every nook and cranny of the ship to find the gold -- each hoping that she will get all nine, and dreading a situation where she finds nothing. How many ways are there to assign the gold bars to the pirates? \\ \hline
\end{tabular}
\caption{Problem 10 presented in five variations. The highlighted text indicates content modifications based on the common version. The mathematical and linguistic obfuscation variations constitute the new text.}
\label{table:problem_forms}
\end{table*}

\section{Combi-Puzzles: Dataset Construction}

\paragraph{}
We believe that combinatorial problems are highly suited for evaluating the mathematical reasoning of LLMs: correct answers are usually expressed as combinatorial formulae or simple numeric results, making the process of verifying the answer efficient and reproducible. Formulaic forms of answers to combinatorial problems can include binomials, factorials, and other combinatorial symbolic representations, both of which are accepted in our experiment. An example of a combinatorial problem with the answer in both forms is shown in Table~\ref{tab:puzzle_example}.

Additionally, given a response in the form of a combinatorial formula, it is usually possible to extract the reasoning steps that led to the final answer. Furthermore, we can find combinatorial problems with a wide range of complexity which can be further adjusted via various text manipulations. These characteristics make combinatorial problems an efficient and practical way to assess mathematical reasoning abilities.

We constructed a dataset of 125 problem variants based on 25 combinatorial problems covering permutations (with and without repetition), combinations, the rules of addition and multiplication, and object arrangements with various restrictions. These problems are representative of basic principles about combinatorics for high school curriculum, and spanning from simple to intermediate complexity levels.

We created five variations of each problem in a controlled manner through manual text alterations of the common variation of the problem set. As a concrete example, Table~\ref{table:problem_forms} shows problem 10 in our dataset in all of its variations.

The \emph{Common} variation is widely available in textbooks about combinatorics, mathematical competition proceeding, and online resources like the AOPS website\footnote{\url{https://artofproblemsolving.com/wiki}.}. This format represents how problems are typically presented to participants to explain combinatorial principles and provide them with illustrative examples.

The \emph{Mathematical} variation is presented in the natural language of mathematics, typically found in academic literature. Statements expressed in this form include mathematical technical terms, concepts (e.g., ``sets'', ``permutations'', ``urn model'') and set phrases (e.g., ``draw with/without replacement'').

The \emph{Adversarial} variation is constructed by introducing text that injects information, such as numerical data, to the common form of the problem statement that is not relevant for solving the problem. Typically, up to a sentence is injected to test the solver's ability to identify relevant information.

The \emph{Parameterisation} variation changes the configuration of the common problem by typically increasing parameter values to expand the answer space and make the problem more challenging.

The \emph{Linguistic Obfuscation} variation is constructed by changing the narrative style of the problem statement and turning it into a narration of a fictional story. This problem variation can include additional descriptions, irrelevant numerical data and fresh names. Every instance of this variation is at least 300 characters long. This approach tests the solvers' ability to extract the core of the problem.

The mathematical, adversarial, parameterisation and linguistic obfuscation forms are fresh presentations of the problems that have been created by the authors specifically for this study. These variations are intended to model the diverse ways a problem can be stated. This way, we can engage the ability of humans and LLMs to identify the correct problem-solving strategy for each problem.

We constructed all variations by systematically applying the guidelines highlighted above such that the underlying mathematical core, the information required to solve it, is preserved. To our knowledge, this study is the first to use linguistic obfuscation to evaluate the performance of large language models.

It is our intention to make our collection available to the scientific community.

\begin{table*} 
\centering \small
\begin{tabular}{|l|l|ccccc|c|} 
\hline 
Model & Additional prompt & Comm. & Math. & Advers. & Param. & Ling.obfus. & Overall \\ 
\hline 
GPT-4 & \textit{None} & \textbf{.82} & \textbf{.94} & \textbf{.77} & \textbf{.67} & \textbf{.70} & \textbf{.78} \\ 
LLaMA-2 & \textit{None} & .22 & .22 & .15 & .19 & .11 & .18 \\ 
LLaMA-3.1 & \textit{None} & .54 & .60 & .50 & .42 & .48 & .51 \\ 
Mixtral & \textit{None} & .42 & .38 & .26 & .23 & .23 & .30 \\ 
\hline 
GPT-4 & ``The short and correct answer is'' & .76 & .90 & .76 & .61 & .66 & .74 \\ 
LLaMA-2 & ``The short and correct answer is'' & .19 & .17 & .05 & .08 & .11 & .12 \\ 
LLaMA-3.1 & ``The short and correct answer is'' & .53 & .61 & .43 & .40 & .52 & .50 \\ 
Mixtral & ``The short and correct answer is'' & .41 & .34 & .18 & .30 & .22 & .29 \\ 
\hline 
\end{tabular} 
\caption{Model averages per variation and overall for two prompting strategies: with (bottom) and without (top) additional prompt added to model input.} 
\label{tab:best_model} 
\end{table*}

\begin{figure*}[t]
    \centering
    \includegraphics[width=1\textwidth]{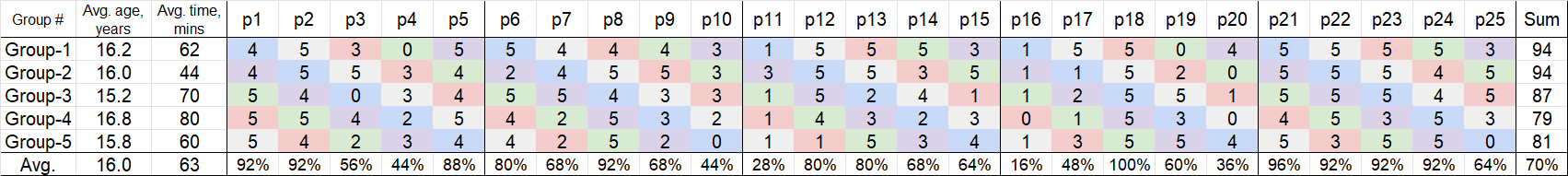}
    \caption{Table represents human groups' scores on the problem set. Each cell represents a problem variation: \hlgrey{common} (grey), \hlblue{mathematical} (blue), \hlred{adversarial} (red), \hlgreen{parameterisation} (green), \hlpurple{linguistic obfuscation} (purple).}
    \label{fig:human_results}
\end{figure*}


\begin{table*}[ht]
\centering \small
\begin{tabular}{|l|ccccc|c|} 
\hline
& Common & Math.\textbf{*} & Advers. & Param. & Ling. obfus. & Overall\textbf{*} \\
\hline
GPT-4 model & .82 & .94 & .77 & .67 & .70 & .78 \\ 
Human students & .78 & .63 & .64 & .70 & .74 & .70 \\
\hline 
P-value & .697 & \textless .05 & .185 & .874 & .714 & \textless .05 \\
\hline
\end{tabular} 
\caption{Average scores for GPT-4 and human students in terms of problems solved correctly across conditions and overall. `*’ represents that condition change was significant ($p < .05$).}
\label{tab:forms_human_v_model} 
\end{table*}

\section{Methods}
\paragraph{}
To answer the three questions under investigation in this paper, we designed two independent experiments, which we executed using our \emph{Combi-Puzzles} dataset: one with LLMs and the other one with humans.

We prompted four models: (i) LLaMA-2-70B-Chat (referred to as ``LLaMA-2'' henceforth), LLaMA-3.1-70B-Instruct (ii) (referred to as `LLaMA-3.1') (iii) Mixtral-8x7B-Intstruct (referred to as ``Mixtral'') and (iv) GPT-4-Turbo (referred to as ``GPT-4'').

For the open-source models LLaMA-2, LLaMA-3.1 and Mixtral we used quantised versions of the models running locally. Access to GPT-4-Turbo is via API calls to the paid service. Model parameters, such as \texttt{temperature} and \texttt{top\_p}, were left unchanged to their default values\footnote{Models specifications and computational budget can be found in the Appendix~\ref{sec:llms-tested}.}. We limited the maximum number of output tokens for all models to MAX\_TOKENS = 2048.

We report statistical significance, where applicable, using the \emph{paired Permutation test } -- a non-parametric test that detects significant differences in the distribution of paired samples.

\subsection{Experiment with LLMs}
\paragraph{}
We used two prompting strategies to elicit solutions to problems in our dataset from the models: (i) no additional prompt (referred to as ``None'') and (ii) one prompt added to the end of the problem statement intended to incentivise the models to give a short answer (referred to as ``The short and correct answer is'' prompt).

For each strategy, we performed $N$ prompting runs per model and per problem variant to account for stochastic variations in model responses. For models LLaMA-2 and Mixtral we set $N=10$, and for models LlaMA-3.1 and GPT-4 we set $N=5$ due to more deterministic behaviour through the generation process.
We recorded the number of questions answered correctly by each model in every run. For each model, we report accuracy averaged across all runs of a particular variation and strategy. Overall accuracy for a model is the average across all problems forms and runs.

We devised guidelines for determining whether a response from a model answers the prompted statement correctly or not. Language models tend to give long, detailed but often inconsistent or contradicting responses. For this purpose, every response was manually assessed by one of the authors against the rules in the guidelines which we summarise here as follows\footnote{Full set of rules is attached in the Appendix~\ref{sec:marking-scheme}.}.

Model responses are considered \emph{correct} if the following conditions hold: (i) the response the response short and correct (no reasoning steps are provided); (ii) contains an explanation with correct calculations leading to a correct answer; (iii) the answer provided is clearly marked as an approximation (for the parameterisation variation problems); (iv) the answer is a combinatorial formula with correct values inserted but with no further calculations.

Model responses are considered \emph{incorrect} if they exhibit one or more of the following characteristics: (i) the response is ambiguous; (ii) is provided in the wrong format (e.g., as programming code); (iii) the answer is empty or incomplete.

We use the permutation test to assess whether alterations to the text have a significant impact on performance and to compare the models against each other.

\subsection{Experiment with Human participants}
\paragraph{}
We invited Ukrainian pupils and undergraduates, who are familiar with the topic of combinatorics and have prior experience in mathematical competitions, to participate in our study. A total of 35 participants, aged 13 to 18 years, agreed to participate in the study.

We organized 35 participants and 125 problem statements into a Latin Square Design. We formed 5 groups of 7 participants, ensuring each group had a roughly similar mean age, and assigned each group one of five distinct problem packs. Each pack contained 25 problems with evenly distributed variations. Our problems were translated from English to Ukrainian so that our participants could reliably read and comprehend the problems. Participants worked individually within their groups.

We applied the same process for evaluating correctness across all responses received by our participants. Each participant provides answers for their problem pack in a short combinatorial or numerical form, which receives a score of 0 (incorrect) or 1 (correct). In case participants do not produce any answer no answer is recorded and the student receives a score of 0 for that problem. The correctness of responses was manually evaluated by one of the authors. Given the diverse formulations a correct response might take (e.g., in the form of the combinatorial formula), in some cases we also had to apply a degree of calculation and logic in our assessment.

In our comparison of LLM versus human performance, we use the best-performing model to represent LLMs, and in the case of human problem solvers, we use a group of the top 5 participants per human group as a representation of human experts.
The distribution of participant groups and problem variants according to all factors along with average age and time spent to return answers is shown in Figure~\ref{fig:human_results}.


\begin{table*}
\centering \small
\begin{tabular}{|l|c|ccccc|}
\hline
& & Common & Math. & Advers. & Param. & Ling. obfus. \\
\hline
GPT-4 model & & .82 & .94 & .77 & .67 & .70 \\
\hline
Common & .82 & - &  &  &  &  \\
Math. & .94 & .12 & - &  &  &  \\
Adversarial & .77 & -.05 & \textbf{-.17*} (.026) & - &  &  \\
Parameterisation & .67 & -.15 & \textbf{-.27*} (.002) & -.10 & - &  \\
Linguistic obfuscation & .70 & -.12 & \textbf{-.24*} (.006) & -.07 & .03 & - \\
\hline
\end{tabular}
\caption{Comparison between variations differences for GPT-4 results across variations. ‘*’ represents that variation difference was significant ($p < .05$).}
\label{tab:forms_diff}
\end{table*}

\section{Experimental Results and Discussion}
We begin our discussion by looking at the results of our first experiment -- 
comparing the mathematical reasoning abilities of the LLMs of interest.
Table~\ref{tab:best_model} shows the performance of each model across variations, with/without additional prompts, and overall. 

From Table~\ref{tab:best_model} we observe that GPT-4 performed best across all problem forms regardless of the prompting strategy. The best overall performance numerically is obtained by issuing questions to GPT-4 without additional prompts (78\% score overall). We did not find significant differences between prompting strategies but due to the differences in performance being more pronounced with the no prompt strategy on average, we will focus on results obtained using this prompting strategy going forward.

We now turn our attention to our second experiment, examining the performance of participants on the \emph{Combi-Puzzles} dataset. As shown in Figure~\ref{fig:human_results}, the participants, who had an average age of 16 years, spent an average of 63 minutes solving the problems and achieved an average solution rate of 70\%.

Collectively, the responses from all participants ensured that every problem was resolved at least once in one of the forms. From Figure~\ref{fig:human_results}, however, we observe that problems 3, 4, 10, 11, 16, 17, 19, and 20 have lower average solution rates than the expected 70\% at 56\%, 44\%, 44\%, 28\%, 16\%, 48\%, 60\% and 36\% solution rate, respectively. 

In a few cases this lower than expected solution rate can be attributed to the complexity of some of the problems. For example, few correct answers have been elicited by participants for problems 11 and 16 (lower than 30\% across all variations). This pattern might be indicative of problems that are difficult to solve for our participants.

In the case of problems 4, 10, 16, 20 and 25 we also observed that none of our participants were able to solve the problem in one or two forms. This might be an indication that under certain variations problems become difficult to understand, even by human participants. We look into these cases in the next section.

Next we look at the difference in performance between human students and LLMs across variations and overall.
From  Table~\ref{tab:forms_human_v_model} we observe that the best model (GPT-4) performed significantly better than human students overall.
The table also shows that GPT-4 excelled in the mathematical form with a score of 94\%, while our students scored significantly lower at 63\%.
We hypothesise that GPT-4 performs exceptionally well in the mathematical form because the material for this variation is similar in structure to training text in the mathematical domain likely seen by GPT-4 during training (e.g., mathematical text derived from mathematics textbooks and tutorials).

For adversarial variation, we observe that GPT-4 received a higher score of 77\% compared to human students 64\% (but not statistically significant, $p=.185$). The adversarial variation is derived from the common statement of a problem by adding irrelevant information. This suggests that humans appeared to be more confused by adversarial information in shorter statements than GPT-4.

However, for the linguistic obfuscation form, the difference between models and humans is much lower (70\% and 74\%, respectively) and even slightly better for humans. However, we did not find the difference statistically significant ($p=.714$). 
One hypothesis as to why GPT-4's performance on the linguistic obfuscation variation is reduced relative to other variations is that the model becomes less able to separate relevant from irrelevant information in longer text.

Another notable difference between the adversarial and linguistic obfuscation variations is the linguistic style: the former is more descriptive, whereas the latter adopts a narration style. This might influence the ability of GPT-4 to extract relevant information from unfamiliar mathematical text. In contrast, our students appear to be more reliable in finding the underlying generalisations.

For the parameterisation form, the performance of GPT-4 compared to humans is almost equal (67\% and 70\%, respectively), which could be an indication that the models lack precision when dealing with larger calculations. In some cases, we observed that despite GPT-4 generating correct reasoning steps, it produced incorrect sequences of expressions in the same chain of calculations, resulting in incorrect results.

We now address question 3. From Table~\ref{tab:forms_diff} we observe that the performance of GPT-4 is sensitive to differences in problem statements introduced by the five forms in our dataset. Specifically, we found that GPT-4's performance at solving combinatorial problems is significantly better when the content is expressed in mathematical form (94\%) when compared to the adversarial, linguistic obfuscation and parameterisation forms (77\%, 70\% and 67\%, respectively). GPT-4's sensitivity to alterations in problem statements demonstrated in our dataset indicates that GPT-4 might not be able to generalise effectively without explicit fine-tuning. However, this sensitivity is only currently visible in our dataset and further study is required to confirm this behaviour.

Meanwhile, the difference in the correctness rates between variations is not significant for our participants. This suggests that the ability of participants to solve combinatorial problems, as observed in our experiment, is not affected by variations in problem statements.


\begin{figure*}[t]
    \centering
    \includegraphics[width=0.48\linewidth]{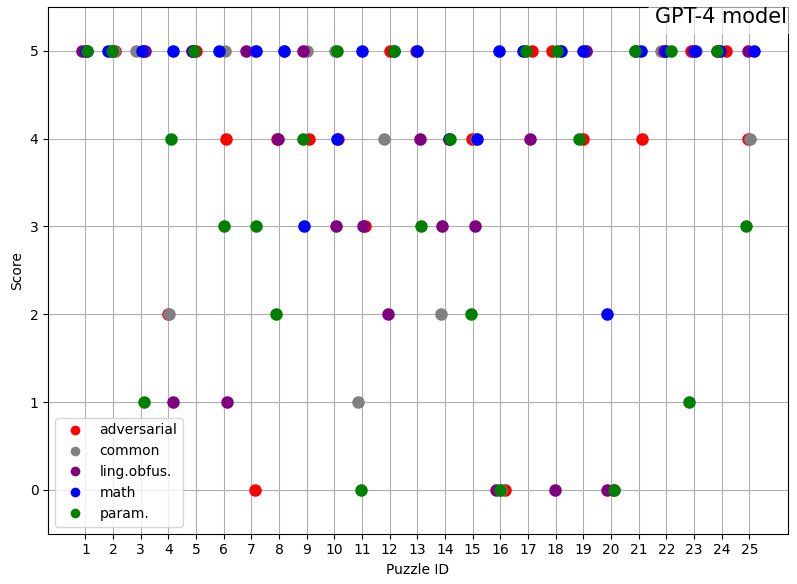} \hfill
    \includegraphics[width=0.48\linewidth]{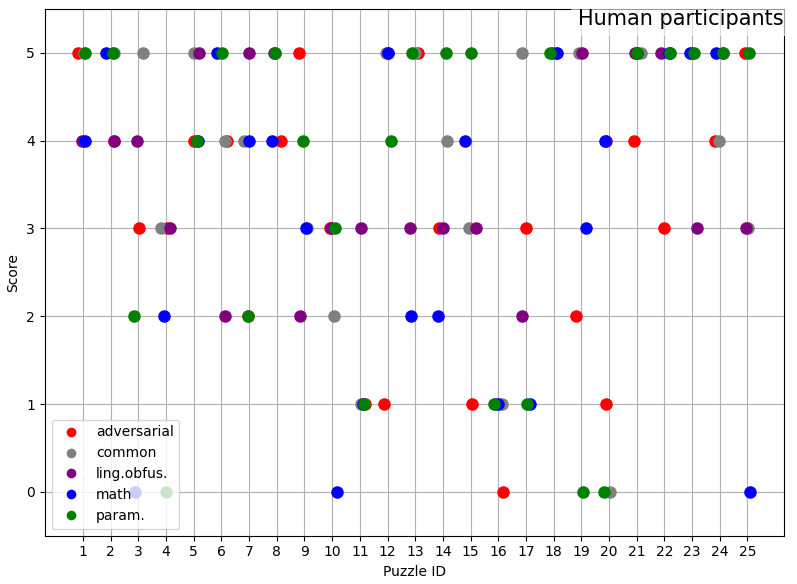}
    \caption{Individual problem scores and the percentage of problems solved correctly are shown for the GPT-4 (left) and human participants (right).}
    \label{fig:humans_v_models_problems}
\end{figure*}

We now focus our investigation on ad-hoc analysis of individual problems. The goal is to identify which problem variations were the most difficult for humans and LLMs, according to our results.

Figure~\ref{fig:humans_v_models_problems} is a visualisation of the number of correct answers provided by GPT-4 and participants for each variation of problems in \emph{Combi-Puzzles}. The x-axis identifies the problem, while the y-axis represents the number of times participants or GPT-4 provided the correct answer. Model scores are out of 5 runs, and human scores are out of 5 participants per group.

Overall, from figure~\ref{fig:humans_v_models_problems} we observe that the number of solutions provided by participants for problems appears to vary more than the number of solutions produced by GPT-4. We observe that the model has a more polarised tendency than participants: if the model is able to solve a problem, it will do so across many variations consistently. In contrast, the ability of human participants to solve a problem in more than one of its forms appears to be dependent upon the individual.

We also observe that in some cases GPT-4 and participants produced zero correct answers for the distinct problems. Specifically, we identified p4/param., p10/math, and p25/math as problems that GPT-4 provided correct solutions on 4, 5 and 5 instances, respectively, to which human participants provided 0 correct solutions. Similarly, we identified problems p18/ling.obfus. and p20/ling.obfus. where our participants provided correct solutions for 5 and 4 instances, respectively, to GPT-4's 0 correct solutions\footnote{Statements of the problems listed here are attached in the Appendix~\ref{sec:problems-poorly-solved}.}.

The combinatorial answer for problem p25/math problem utilizes separately solved independent subcases of the problem with exactly 0 black balls left, 1 black ball left, etc., and then summed up together: $C(25, 0) + C(25, 1) + C(25, 2) + C(25, 3) = \sum_{i=0}^{3} C(3, i)$.
Looking at GPT-4's response, we find that the model was able to provide the correct response showing its ability to divide a problem into its subproblems, solve the subproblems separately, and apply the rule of sum to get to the final answer. This problem is one of the most difficult problems in our dataset and the mathematical variation was not solved by any of our participants.

GPT-4 also performed better than our participants in problem p10/math. We noticed that our participants did not pay attention to the uniqueness of the sets involved -- ``How many \textit{distinct sets} of balls are there?''-- and provided an incorrect combinatorial answer of $3^9$. In contrast, GPT-4 understood the question correctly and applied the right combinatorial formula to get the answer, $C(9+(3-1), 3-1)=C(11,2) = 55$.

In some cases, GPT-4 demonstrated the ability to identify the relevant concepts mentioned in the problem statements and apply them to solve larger instances of the problem. One example of this phenomenon is GPT-4's success at solving a parameterised version of the problem about cinema tickets (p4/param.) by recognizing the concept ``Catalan number sequence''. This enabled the model to skip intermediate reasoning steps and get directly to the final answer by utilising initial numbers for the formula: $C_n = (2n)! / ((n + 1)! n!)$.

In other cases, GPT-4 demonstrated difficulty in solving linguistic obfuscation variations of problems that we consider to be amongst the simplest in our dataset. One such instance is p18/ling.obfus., which GPT-4 solved correctly 0 times. The problem requires counting ``ways to travel between the cities'', in this case: $3 \times 5 = 15$. The addition of ``a choice of 7 horses at Ponyville'' as adversarial information confused GPT-4, which multiplied the final answer by 7, resulting in an incorrect response: $3 \times 5 \times 7 = 105$.

Another problem where human participants performed better than GPT-4 is the p20/ling.obfus. Humans correctly identified the narrative of the problem, which could be achieved by simplifying the statement and transforming the problem into a sequence with constraints, as in the mathematical form of the same problem p20/math:
``How many distinct strings could be created from the string $ABCDE$ if $A$ must go before $B$ and $C$ must go before $D$?''.
The correct answer for the problem is $30$. GPT-4 attempted to list some possible combinations that satisfy the problem statement but failed to include all of them, which then resulted in an incorrect final answer.

\section{Conclusion}
\paragraph{}
We introduced the \emph{Combi-Puzzles} dataset for evaluating the reasoning ability of LLMs in combinatorial problems and a methodology for constructing datasets for this task. 

Our experimental results overwhelmingly point to GPT-4 as the LLM that best answers combinatorial mathematics problems. We found statistically significant evidence suggesting that GPT-4 is better than human participants at producing correct answers. 

Our results suggest that GPT-4 ability to reason peaks when combinatorial problems are expressed in the natural language style of mathematical textbooks (mathematical variation, 94\%). However, we have also seen that in our dataset, GPT-4's performance drops significantly from the aforementioned peak when additional information is injected into the problem statement (adversarial variation, 77\%) and when the problem statement is presented in an unfamiliar narrative style (linguistic obfuscation, 70\%). Although the performance of our participants varied slightly numerically, we did not find a significant difference in the success rate of their responses across variations. 


We plan to make our collection accessible to the scientific community.

In future work, we intend to design experiments specifically for examining the reasoning steps taken by LLMs and human participants in order to better understand their cognitive differences in mathematical reasoning.

\section{Limitations}

We identify three main limitations of our approach. The first concerns the human participants' experience. The age of the participants ranges from 13 to 18 years old, with a mean age of 16. While all participants have experience in mathematical competitions and are familiar with combinatorics, their backgrounds vary due to age differences, which may affect their preparedness for solving the problem variants in their novel forms. We controlled for age differences by distributing them evenly among groups, allowing this limitation to contribute constructively to the study's aims.

The second limitation of our study is the size of the \emph{Combi-Puzzles} dataset; owing to its handcrafted nature, it contains 125 problem variants, which may not fully capture the diversity of scenarios LLMs can encounter. This limits the generalisability of our results to broader applications.

The final limitation is the quality of the translated puzzle statements of our dataset. The problems were presented to the human participants in Ukranian, whereas the original problems were written in English. Although we took care of preserving the same alterations with the text of each problem variant, the translated dataset may slightly deviate from the original form. Specifically, in order to ensure that the Ukrainian translation of statements remained familiar to our participants, we replaced set phrases in English with their Ukrainian equivalents.  For example, the set phrase ``drawing balls with replacement'' was translated directly into ``after you take a ball you put it back''. These adaptations are motivated by differences in mathematical set phrases between Ukranian and English languages as observed in mathematical textbooks.

\section{Ethics Statement}

This study was conducted as a controlled experiment approved by the university's Ethics Committee. We recruited 35 current and former students from the Ukrainian STEM project Kvanta\footnote{\url{https://kvanta.xyz/}}, ensuring informed consent from participants and parental consent for younger participants, who are below 16 years old. The experiment assessed mathematical reasoning skills online, with data collected anonymously. Participation was voluntary, and participants were aware of their right to withdraw at any time. The guidelines included general information about the number of problems presented, their complexity and variance in style. No monetary compensation was provided, but involvement was offered as an educational opportunity. Communication was clear, emphasising that the study is separate from regular project activities.

\bibliography{references}

\appendix
\section{Marking Scheme}
\label{sec:marking-scheme}
We provide the Marking Scheme -- a list of rules we applied to give scores for models' outputs. In some cases, humans and LLMs return responses not in numerical form but as a combinatorial formula with the use of binomials, factorials, and other combinatorial symbolic representations, e.g. \(C(11, 2) = 55\) -- such responses are also accepted if they are equal to the correct answer numerically. We list all the rules applied for the marking scheme, including the "grey area" cases. The Marking Scheme can be found below.\\

\textbf{Criteria for Score 1 (Correct):}
\begin{enumerate}
    \item \textit{Score: 1} - Answer is short and correct.
    \item \textit{Score: 1} - There is an indicated correct answer at the beginning, even if incorrect or irrelevant information follows.
    \item \textit{Score: 1} - There is an indicated correct answer at the end, even if incorrect or irrelevant information precedes it.
    \item \textit{Score: 1} - Model lists several answers and indicates the correct one.
    \item \textit{Score: 1} - Model starts with an incorrect answer but then reasons to a correct one.
    \item \textit{Score: 1} - Combinatorial answer is correct, but a model provides an approximation for a numerical answer (e.g., using words "approximately", "about", "\(\approx\)", etc.).
\end{enumerate}

\textbf{Criteria for Score 0 (Incorrect):}
\begin{enumerate}
    \item \textit{Score: 0} - Answer is incorrect.
    \item \textit{Score: 0} - Answer is empty.
    \item \textit{Score: 0} - Response lacks numerical or combinatorial parts.
    \item \textit{Score: 0} - Model hasn't finished a solution with a final answer.
    \item \textit{Score: 0} - Model lists several answers (possibly including the correct answer) and indicates the wrong one or none.
    \item \textit{Score: 0} - Response is a code written in a programming language.
    \item \textit{Score: 0} - Model has correct reasoning steps but concludes with a wrong answer.
    \item \textit{Score: 0} - Incorrect equality exists, connecting multiple answers with one being correct; answers connected with an equal sign or words like "or", "equals", etc.
    \item \textit{Score: 0} - The answer is ambiguous (no final response is given).
\end{enumerate}

 
\begin{table*}[ht]
\centering
\begin{tabular}{|p{2.5cm}|p{1.2cm}|p{1cm}|p{1.5cm}|p{1.5cm}|p{1.6cm}|p{2cm}|p{1.5cm}|}
\hline
Model & Params & Is quant. & Q. method & Context length & Knowledge cutoff & Access & Model Creator \\
\hline
LLaMA-2-70B-Chat & 69B & \centering \ding{51} & Q4\_K\_M & 4k & Jul 2023 & Open-source & Meta \\
\hline
LLaMA-3.1-70B-Instruct & 70.6B & \centering \ding{51} & Q4\_K\_M & 128k & Dec 2023 & Open-source & Meta \\
\hline
Mixtral-7Bx8-Instruct & 46.7B & \centering \ding{51} & Q4\_K\_M & 32k & N/A & Open-source & Mistral AI \\
\hline
GPT-4-Turbo-preview-v1106 & N/A & \centering \ding{55} & N/A & 128k & Apr 2023 & API & OpenAI \\
\hline
\end{tabular}
\caption{Specifications of models used in the experiment.}
\label{tab:model_card}
\end{table*}

\section{Problems Poorly Solved}
\label{sec:problems-poorly-solved}
This appendix contains instances of problems that received the lowest scores from human participants or LLMs. We identify each problem and its variation in the format ProblemID/variation. The bracketed information next to the problem identifier indicates how many times the problem was correctly solved by (i) GPT-4 and (ii) participants, in that order. 
\begin{enumerate}
\item \textbf{p4/param.} (Model-Humans 4:0) \\
There is a line in a cinema for a movie. The ticket costs 5 pounds. 5 people have a 10 pound banknote each and 5 people have a 5 pound banknote each. At the beginning the ticket cashier does not have any change. How many distinct ways are there to form a line, so that everybody can buy a ticket?

\item \textbf{p10/math.} (Model-Humans 5:0) \\
In an urn there are 3 balls numbered 1 to 3. You draw 9 times with replacement. How many distinct sets of balls are there?

\item \textbf{p18/ling.obfus.} (Model-Humans 0:5) \\
A knight, who lives in Greenfields town, intends to participate at the annual jousting event. First, he has to go Ponyville town, where he will get his noble horse. There are 3 roads between Greenfields and Ponyville. He has a choice of 7 horses at Ponyville. After that, he has to go to Saddleford town and pick the brand-new shiny and comfortable saddle. There are 5 roads between Ponyville and Saddleford. How many ways are there to travel from Greenfields to Saddleford through Ponyville?

\item \textbf{p20/ling.obfus.} (Model-Humans 0:4) \\
You oversee two islands and a ceremonial boat, and your job is to make sure several preparations for an important festival are made. On each island, the highest tree needs to be felled and then carved into a totem to be placed on the beach. As soon as one of the two tree is felled, a telegram is transmitted to the mainland and given a time stamp. When one of the two totems is placed on the beach, another telegram is transmitted. You also need to make sure the ceremonial boat is decorated. Again, when this task is finished, a telegram is transmitted. You have workers on each island, and on the boat. You do not know how long each task takes. At the end of the preparations, you look a the order of the incoming telegrams. How many different sequences are there in which the telegrams can come in?

\item \textbf{p25/math.} (Model-Humans 5:0) \\
There is an urn with 25 white balls numbered 1 to 25 and 25 black balls numbered 1 to 25. You draw 25 times without replacement. Every time you draw a ball with label N, other ball with the same label but other color disappears from the urn. How many distinct sets of balls with 3 or less black balls are there?
\end{enumerate}

\section{Large Language Models Tested}
\label{sec:llms-tested}
The specifications of the large language models used in our experiments are detailed in Table \ref{tab:model_card}. The open-source models, which are provided in the GGUF format -- a binary format optimized for quick loading and saving to enhance the efficiency of the inference process -- have been enhanced further by applying K-means Quantization through the \texttt{llama.cpp} library. Within the \texttt{llama.cpp} context, Q4\_K\_M denotes a specific quantization method, whereby `Q' stands for Quantization, `4' signifies the use of four bits in the quantization process, `K' refers to k-means clustering, and `M' indicates the model's size post-quantization, classified as Small (S), Medium (M), or Large (L).

The open-source models were run locally on Nvidia Quadro RTX 8000 (48GB of RAM). Each problem variant has been processed for 1-2 minutes depending on the model, prompting strategy and problem variant, totalling up to 20 GPU hours per model for $N = 5$ runs for all 125 problem variants from the dataset. The GPT-4 model was accessed via API online and has been processed in a few hours, specifications of the system configurations are unknown.

The tested models are available at:

\begin{itemize}
    \item \textbf{LLaMA-2-70B-Chat}:\\
    \url{https://huggingface.co/TheBloke/Llama-2-70B-Chat-GGUF}.
    \item \textbf{LLaMA-3.1-70B-Instruct}:\\
    \url{https://huggingface.co/bartowski/Meta-Llama-3.1-70B-Instruct-GGUF}.
    \item \textbf{Mixtral-8x7B-Instruct}:\\
    \url{https://huggingface.co/TheBloke/Mixtral-8x7B-Instruct-v0.1-GGUF}.
    \item \textbf{GPT-4-Turbo-preview-v1106}: (accessible via API)\\ 
    \url{https://platform.openai.com/docs/models/#gpt-4-turbo-and-gpt-4}.
\end{itemize}

\end{document}